\documentclass[11pt]{article}
\usepackage{jair}
\usepackage{named}
\usepackage{tabularx}
\usepackage{xparse}
\usepackage{times}
\usepackage{helvet}
\usepackage{courier}
\usepackage{amsmath}
\usepackage{amsthm}
\usepackage{amssymb}
\usepackage{xspace}
\usepackage{relsize}
\usepackage{aaai_my}
\usepackage{graphicx}
\usepackage{abbrev}
\usepackage{multirow}
\usepackage{comment}
\usepackage[normalem]{ulem}
\usepackage[frozencache]{minted}
\usepackage{algorithm}
\usepackage[noend]{algorithmic}
\usepackage{url}

\usepackage{fancyhdr}
\fancypagestyle{plain}{%
\fancyhf{} 
\fancyfoot[C]{\iffloatpage{}{\thepage}} 

}
\usepackage{longtable}
\makeatletter
\renewcommand{\@todons}[2]{%
  \uline{#1}%
  \marginpar[%
    \hfill\relsize{-1}%
    \protect{%
      \rotatebox[origin=c]{90}{\bf todo: #2}}%
    $\triangleright$]{%
    \relsize{-1} $\triangleleft$%
    \protect{%
      \rotatebox[origin=c]{-90}{\bf todo: #2}}}}
\renewcommand{\@todos}[2]{%
  \uline{#1}%
  \marginpar[%
    \hfill\relsize{-1}%
    \protect{%
      \rotatebox[origin=c]{90}{done \it #2}}%
    $\triangleright$]{%
    \relsize{-1} $\triangleleft$%
    \protect{%
      \rotatebox[origin=c]{-90}{done \it #2}}}}
\makeatother

\usepackage{array}
\newcolumntype{H}{>{\setbox0=\hbox\bgroup}c<{\egroup}@{}}
\newcolumntype{E}{r!{$\pm$}l}

\ShortHeadings{Classical Planning in Deep Latent Space (Accepted in AAAI-18, Extended Manuscript)}{Asai \& Fukunaga (Accepted in AAAI-18, Extended Manuscript)}
\firstpageno{1}
\author{
\name Masataro Asai \email guicho2.71828\textcircled{$\alpha$}gmail.com\\
\name Alex Fukunaga \\
\addr Graduate School of Arts and Sciences, University of Tokyo}

\title{Classical Planning in Deep Latent Space:\\ Bridging the Subsymbolic-Symbolic Boundary} 

\begin{document}
\maketitle

\begin{hidden}

* too much ``''s make the sentence look scattered and visually less recognizable. ``e.g.'' also.

* \em, \bf, \it are all obsolete \TeX primitives, and it does not take effect properly --- for example, {\bf {\it aaa}} shows ``aaa'' in italic but NOT IN BOLD. Use \emph{}, \textit{}, \textbf{} and so on.

* always use \ff, \fd, \cea, \pr, \mv , and do not use it directly, e.g. FF, FD/LAMA2011, etc. 

* use of footnotes should be minimized.

* IPC2011 should always be \ipc . The definition can later be modified in abbrev.sty .

* prefer separated words over hyphened words. domain
  independent>domain-independent, planner independent >
  planner-independent.

* Table, Figure, Fig., should not be used directly. Always use \refig and \reftbl. When the development flag is enabled, direct use of \ref signals an error.

* Caption ends with a period.
\end{hidden}


\begin{abstract}
Current domain-independent, classical planners require symbolic models of the problem domain and instance as input, resulting in a knowledge acquisition bottleneck.
Meanwhile, although deep learning has achieved significant success in many fields, the knowledge is encoded in a subsymbolic representation which is incompatible with symbolic systems such as planners.
We propose \latentplanner, an unsupervised architecture combining deep learning and classical planning.
Given only an unlabeled set of image pairs showing a subset of transitions allowed in the environment (training inputs),
and a pair of images representing the initial and the goal states (planning inputs),
\latentplanner finds a plan to the goal state in a symbolic latent space and returns a visualized plan execution.
The contribution of this paper is twofold:
(1) State Autoencoder, which finds a propositional state representation of the environment using a Variational Autoencoder.
It generates a discrete latent vector from the images, based on which a PDDL model can be constructed and then solved by an off-the-shelf planner.
(2) Action Autoencoder / Discriminator, a neural architecture which jointly finds the action symbols and the implicit action models (preconditions/effects),
and provides a successor function for the implicit graph search.
We evaluate \latentplanner using image-based versions of 3 planning domains: 8-puzzle, Towers of Hanoi and LightsOut.
\end{abstract}

\section*{Note}

\textbf{This is an extended manuscript of the paper accepted in AAAI-18.
The contents of AAAI-18 submission itself is significantly extended from what has been published in
Arxiv, KEPS-17, NeSy-17 or Cognitum-17 workshops.
Over half of the paper describing (2) is new.
Additionally, this manuscript contains the contents in the supplemental material of AAAI-18 submission.
These implementation/experimental details are moved to the Appendix.
}

\subsection*{Note to the ML / deep learning researchers}

This article
combines the Machine Learning systems and the classical, logic-based symbolic systems.
Some readers may not be familiar with NNs and related fields like you are, thus
we include very basic description of the architectures and the training methods.

\section{Introduction}

\label{sec:introduction}

Recent advances in domain-independent planning have greatly enhanced their capabilities.
However, planning problems need to be provided to the planner in a structured, symbolic representation such as PDDL \cite{McDermott00}, and in general, such symbolic models need to be provided by a human, either directly in a modeling language such as PDDL, or via a compiler which transforms some other symbolic problem representation into PDDL.
This results in the {\it knowledge-acquisition bottleneck}, where the modeling step is sometimes the bottleneck in the problem-solving cycle.
In addition, the requirement for symbolic input poses a significant obstacle to applying planning in {\it new, unforeseen} situations where no human is available to create such a model or a generator, e.g., autonomous spacecraft exploration.
In particular this first requires generating symbols from raw sensor input, i.e., the {\it symbol grounding problem} \cite{Steels2008}.

Recently,  significant advances have been made in neural network (NN) deep learning approaches for perceptually-based cognitive tasks including image classification \cite{deng2009imagenet}, object recognition \cite{ren2015faster}, speech recognition \cite{deng2013new}, machine translation
as well as  NN-based problem-solving systems \cite{dqn,neuraltm}.
However, the current state-of-the-art, pure NN-based systems do not yet provide guarantees provided by symbolic planning systems, such as deterministic completeness and solution optimality.


Using a NN-based perceptual system to 
{\it automatically} provide input models for domain-independent planners could greatly expand the applicability of planning technology and offer the benefits of both paradigms.
\emph{We consider the problem of robustly,  automatically bridging the gap between such subsymbolic representations and the symbolic representations required by domain-independent planners}.

\begin{figure}[tbp]
 \centering
 \includegraphics[width=\linewidth]{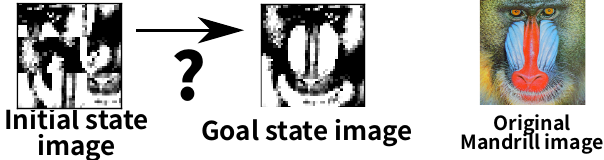}
 \caption{An image-based 8-puzzle.}
 \label{fig:mandrill-intro}
\end{figure}

\refig{fig:mandrill-intro} (left) shows a scrambled, 3x3 tiled version of the photograph on the right, i.e., an image-based instance of the 8-puzzle.
Even for humans, this photograph-based task is arguably more difficult to solve than the standard 8-puzzle because of the distracting visual aspects.
We seek a domain-independent system which, given only a set of unlabeled images showing the valid moves for this image-based puzzle, finds an optimal solution to the puzzle.
Although the 8-puzzle is trivial for symbolic planners, solving this image-based problem with 
a domain-independent system which (1)  \emph{has no prior assumptions/knowledge}
 (e.g., ``sliding objects'', ``tile arrangement''), and (2) \emph{must acquire all knowledge from the images}, is nontrivial.
Such a system should not make assumptions about the image (e.g., ``a grid-like structure'').
The only assumption allowed about the nature of the task is that it can be modeled as a classical planning problem (deterministic and fully observable).

We propose Latent-space Planner (\latentplanner), an architecture which completely automatically generates a 
symbolic problem representation from the subsymbolic input, which can be used as the input for a classical planner.
\latentplanner consists of 3 components: (1) a NN-based {\it State Autoencoder} (SAE), which provides a bidirectional mapping between the raw images of the environment and its propositional representation, 
(2) an {\it action model acquisition} (AMA) system which grounds the action symbols and learns the action model,
and (3) a symbolic planner. 
Given only a set of {\it unlabeled images} of the environment, and in an unsupervised manner,
we train the SAE and AMA to generate its symbolic representation.
Then, given a planning problem instance as a pair of initial and goal images such as \refig{fig:mandrill-intro}, \latentplanner 
uses the SAE to map the problem to a symbolic planning instance, invokes a planner, then visualizes the plan execution.
We evaluate \latentplanner using image-based versions of the 8-puzzle, LightsOut, and Towers of Hanoi domains.

\section{Background}
\label{sec:background}

%

\subsection{Classical Planning}

Classical Planning is achieving a significant advance in the recent years due to the success of heuristic search.
The input problem to a Classical Planning solver (a \emph{planner}) is a 5-tuple 
$\Pi=\brackets{P,O,I,G,A}$ where $P$ defines a set of first-order predicates, $O$ is a set of symbols called \emph{objects}, $I$ is the initial state, $G$ is a set of goal conditions, and $A$ is a set of actions which defines the state transitions in the search space.
A state is an assignment of boolean values to the set of propositional variables, while a condition is a partial assignment that assigns values only to a subset of propositions.
Each proposition is an instantiation of a predicate with objects.
Lifted action schema $a \in A$ is a 5-tuple
$\brackets{\mbox{\textit{params}}, \mbox{\textit{pre}}, e^+, e^-, c}$
where each element means the set of parameters, preconditions,
add-effects, delete-effects and the cost, respectively.
Parameter substitution using objects in $O$ instantiates \emph{ground actions}.
When $c$ is not specified, it is usually assumed $c=1$.
These inputs are described in a PDDL modeling language \cite{bacchus2000} and its extensions.

\begin{figure}[htb]
\centering
\begin{minipage}[c]{0.2\linewidth}
 \begin{align*}
        & Empty(x_0, y_0)          \\
  \land & At   (x_1, y_0, panel_6) \\
  \land & Up   (y_0, y_1)          \\
  \land & Down (y_1, y_0)          \\
  \land & Right(x_0, x_1)          \\
  \land & Left (x_1, x_0)       \ldots 
 \end{align*}
\end{minipage}
\begin{minipage}[c]{0.4\linewidth}
 \begin{minted}[]{common-lisp}
  (empty x0 y0)
  (at    x1 y0 panel6)
  (up    y0 y1)
  (down  y1 y0)
  (right x0 x1)
  (left  x1 x0)...
 \end{minted}
\end{minipage}
\begin{minipage}[c]{0.3\linewidth}
 \includegraphics{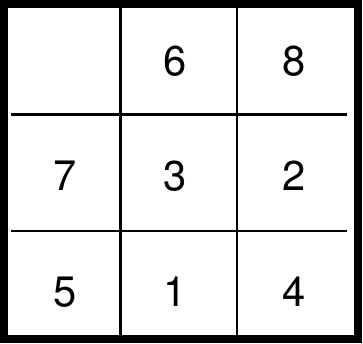}
\end{minipage}
\caption{One possible state representation of a
3x3 sliding tile puzzle (8-puzzle) in the first order logic formula and its
corresponding PDDL notation. It contains predicate symbols 
\pddl{empty}, \pddl{up}, \pddl{down}, \pddl{left}, \pddl{right}, \pddl{at} as well as
object symbols such as \pddl{x}$_i$, \pddl{y}$_i$, \pddl{panel}$_j$ for $i \in \braces{0..3}$ and $j\in \braces{1..8}$.
}
\label{8puzzle-pddl}
\end{figure}

\refig{8puzzle-pddl} shows one possible representation of a state in
3x3 sliding tile puzzle (8-puzzle) in the First Order Logic formula, and the
representation of the same state using PDDL.

\begin{figure}[htb]
\begin{minipage}[c]{0.19\linewidth}
 When $Empty(x, y_{old}) \land at(x, y_{new}, p) \land up(y_{new}, y_{old})$ ;
 
 then $\lnot Empty(x,y_{old}) \land Empty(x,y_{new}) \land \lnot at(x, y_{new}, p) \land at(x, y_{old}, p)$
\end{minipage}
\begin{minipage}[c]{0.56\linewidth}
 \begin{minted}[]{common-lisp}
 ;; Translates to a PDDL model below:
 (:action slide-up ...
  :precondition
  (and (empty ?x ?y-old)
       (at ?x ?y-new ?p) ...)
  :effects
  (and (not (empty ?x ?y-old))
       (empty ?x ?y-new)
       (not (at ?x ?y-new ?p))
       (at ?x ?y-old ?p)))
 \end{minted}
\end{minipage}
\begin{minipage}[c]{0.24\linewidth}
 \includegraphics{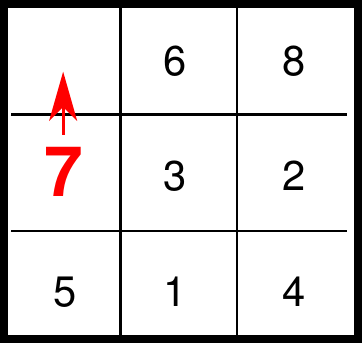}
\end{minipage}
\caption{One possible action representation of sliding up a tile in 3x3
sliding tile puzzle in (left) the first order logic formula and (middle)
its corresponding PDDL notation. In addition to
\refig{8puzzle-pddl}, it further contains an action symbol
\pddl{slide-up}.  } \label{8puzzle-action-pddl}
\end{figure}

The task of a planning problem is to find a path from the initial state
$I$ to some goal state $s^*\supseteq G$, using the state transition
rules in $A$. A state $s$ can be transformed into a new state $t$ by
applying a ground action $a$ when $s\supseteq \mbox{\textit{pre}}$, and
then $t=(s\setminus e^-)\cup e^+$ \cite{bacchus2000}. This transition
can also be viewed as applying a state transition function $a$ to $s$,
which can be written as $t=a(s)$.


\sota planners solve this problem as a path finding problem on a
implicit graph defined by the state transition rules. They usually
employ forward state space heuristic search, such as \astar (for finding
the shortest path) or Greedy Best-First Search (for finding a suboptimal
path more quickly). Thanks to the variety of successful
domain-independent heuristic functions
\cite{Helmert2009,sievers2012efficient,helmert2007flexible,bonet2013admissible,hoffmann01,Helmert04,richter2008landmarks},
current \lsota planners can scale to larger problems which requires to
find a plan consisting of more than 1000 steps \cite{Asai2015}.


\subsection{Knowledge Acquisition Bottleneck}

While ideally, symbolic models like \refig{8puzzle-pddl} should be learned/generated by the machine itself,
in practice, they must be hand-coded by a human, resulting in the 
%
so-called Knowledge Acquisition Bottleneck \cite{cullen88}, 
which refers to the excessive cost of human involvement in converting real-world problems into inputs for symbolic AI systems.

In order to fully automatically acquire symbolic models for Classical Planning, 
\textbf{Symbol Grounding} and \textbf{Action Model Acquisition} (AMA) are necessary.
\textbf{Symbol Grounding} is an unsupervised process of establishing a mapping
from huge, noisy, continuous, unstructured inputs
to a set of compact, 
discrete, identifiable (structured) entities, i.e., symbols.
For example, PDDL has six kinds of symbols: Objects, predicates, propositions, actions, problems and domains (\reftbl{tab:type-of-symbols}).
Each type of symbol requires its own mechanism for grounding.
For example, the large body of work in the image processing community on recognizing 
objects (e.g. faces) and their attributes (male, female) in images, or scenes in videos (e.g. cooking)
can be viewed as corresponding to grounding the object, predicate and action symbols, respectively.

\begin{table}[tbp] 
\centering
\begin{tabular}{ll}
Types of symbols & \\
\hline
Object symbols    & \textbf{panel7, x\(_{\text{0}}\), y\(_{\text{0}}\)} \ldots{}               \\
Predicate symbols & (\textbf{empty} ?x ?y) (\textbf{up} ?y\(_{\text{0}}\) ?y\(_{\text{1}}\))   \\
Propositions      & \textbf{empty\(_{\text{5}}\)} = (empty x\(_{\text{2}}\) y\(_{\text{1}}\)) (6th application) \\
Action symbols    & (\textbf{slide-up} panel\(_{\text{7}}\) x\(_{\text{0}}\) y\(_{\text{1}}\)) \\
Problem symbols   & \textbf{eight-puzzle-instance1504}, etc.                                   \\
Domain  symbols   & \textbf{eight-puzzle}, \textbf{hanoi}                                      \\
\hline
\end{tabular}
\caption{6 types of symbols in a PDDL definition.}
\label{tab:type-of-symbols}
\end{table}

In contrast, an \textbf{Action Model} is a 
symbolic/subsymbolic data structure representing the causality in the transitions of the world,
which, in PDDL, consists of preconditions and effects (\refig{8puzzle-pddl}). 
In this paper, we focus on propositional and action symbols, as well as AMA, leaving first-order symbols (predicates, objects) as future work.

\subsection{Action Model Acquisition (AMA) Methods}

Existing methods require symbolic
or near-symbolic, structured inputs. ARMS \cite{YangWJ07}, LOCM
\cite{CresswellMW13}, and \citeauthor{MouraoZPS12} (\citeyear{MouraoZPS12})
assume the action, object, predicate symbols.
Framer \cite{lindsay2017framer} parses natural language texts and emits PDDL,
but requires a clear grammatical structure and word consistency. 

\citeauthor{KonidarisKL14} generated PDDL from
semi-MDP (\citeyear{KonidarisKL14}).
%
They convert a probabilistic \emph{model} into a propositional \emph{model},
i.e., they do not generate a model from unstructured inputs.
In fact, options ($\approx$ actions) in their semi-MDP have names assigned by a human (\pddl{move}/\pddl{interact}),
and state variables are identifiable entities
(x/y distances toward objects, light level, state of a switch) i.e. already symbolic.


Previous work in Learning from Observation, which could take images (unstructured input),
typically assume domain-dependent hand-coded symbol extractors,
such as \emph{ellipse detectors} for tic-tac-toe board data
which immediately obtains propositions \cite{BarbuNS10}.
\citeauthor{Kaiser12} (\citeyear{Kaiser12}) similarly assumes grids and pieces
to obtain the relational structures in the board image.


\subsection{Autoencoders and Latent Representations}
An Autoencoder (AE) is a type of feed-forward neural network that learns an identity function whose output matches the input \cite{hinton2006reducing}.
Its intermediate layer (typically smaller than the input) has a compressed, \emph{latent representation} of the input.
AEs are trained by backpropagation (BP) to minimize the reconstruction loss, the distance between the input and the output according to a distance function such as Euclidean distance.
NNs, including AEs, typically have continuous activations and integrating them with propositional reasoners is not straightforward.


\section{\latentplanner:  System Architecture}
\label{sec:overview}

\begin{figure}[tbp]
 \centering
 \includegraphics[width=\linewidth]{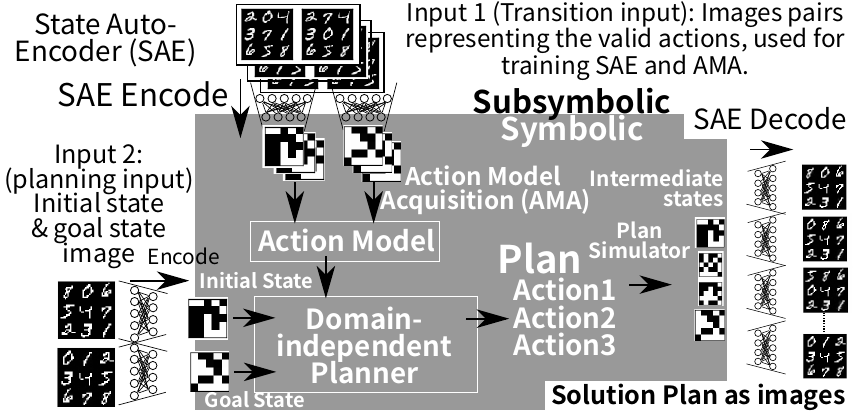}
 \caption{Classical planning in latent space:
We use the learned State Autoencoder (\refsec{sec:state-autoencoder}) to convert pairs of images $(\before,\after)$ first to symbolic transitions, from which the AMA component generates an action model.
We also encode the initial and goal state images into symbolic initial/goal states.
A classical planner finds the symbolic solution plan.
Finally, intermediate states in the plan are decoded back to a human-comprehensible image sequence.}
\label{fig:overview}
\end{figure}

This section describes the high-level architecture of \latentplanner (\refig{fig:overview}).
\latentplanner takes two inputs.
The first input is the \emph{transition input} $Tr$, a set of pairs of raw data.
Each pair $tr_i=(\before_i, \after_i) \in Tr$ represents a transition of the environment before and after some action is executed.
The second input is the \emph{planning input} $(i, g)$, a pair of raw data, which corresponds to the initial and the goal state of the environment.
The output of \latentplanner is a data sequence representing the plan execution that reaches $g$ from $i$.
While we present an image-based implementation (``data'' = raw images),
the architecture itself does not make such assumptions
and could be applied to the other data formats e.g. audio/text.


\latentplanner works in 3 phases.
In Phase 1, a \emph{State Autoencoder} (SAE) learns a bidirectional mapping between raw data (e.g., images)
 and propositional states from a set of unlabeled, random snapshots of the environment.
The $Encode$ function maps images to propositional states, and $Decode$ function maps the propositional states back to images.
After training the SAE from $\braces{\before_i, \after_i\ldots}$,
we apply $Encode$ to each $tr_i \in Tr$ and obtain $(Encode(\before_i),$ $Encode(\after_i))=$ $(s_i,t_i)=$ $\overline{tr}_i\in \overline{Tr}$,
the symbolic representations (latent space vectors) of the transitions.



In Phase 2, an AMA method identifies the action symbols from $\overline{Tr}$ and learns an action model, both in an unsupervised manner.
We propose two approaches: AMA$_1$ directly generates a PDDL and AMA$_2$ produces a successor function (implicit model).
Both methods have advantages and drawbacks.
AMA$_1$ is a trivial AMA method designed to show the feasibility of SAE-generated propositional symbols. It does not learn/generalize from examples, instead requiring all valid state transitions. However, since AMA$_1$ directly produces a PDDL model, it serves as a demonstration that in principle, the approach is compatible with existing planners.
AMA$_2$ is a novel NN architecture which jointly learns action symbols and action models from a small subset of transitions in an unsupervised manner. Unlike existing methods, AMA$_2$ does not require action symbols. Since it does not produce PDDL, it needs a search algorithm (such as A*) for AMA$_2$, or semi-declarative symbolic planners \cite{frances2017purely}, instead of PDDL-based solvers.

In Phase 3, a planning problem instance is generated from the planning input $(i,g)$.
These are converted to symbolic states by the SAE, and the symbolic planner solves the problem.
For example, an 8-puzzle problem instance consists of an image of the start (scrambled) configuration of the puzzle ($i$), and an image of the solved state ($g$).

Since the intermediate states comprising the plan are SAE-generated latent bit vectors, the ``meaning'' of each state (and thus the plan) is not necessarily clear to a human observer.
However, in the final step, we obtain a step-by-step visualization of the plan execution (e.g. \refig{fig:mnist})
by $Decode$'ing the latent bit vectors for each intermediate state.

In this paper, we evaluate \latentplanner as a high-level planner using puzzle domains such as the 8-puzzle.
Mapping a high-level action to low-level actuation sequences via a motion planner is beyond the scope of this paper.


\section{SAE as a Gumbel-Softmax VAE}
\label{sec:state-autoencoder}

First, note that a direct 1-to-1 mapping from images to propositions can be trivially obtained from
the array of discretized pixel values or an image hash function.
However, such a trivial SAE lacks the crucial properties of
\emph{generalization} -- ability to describe unseen world states with the same symbols --
\emph{robustness} -- two similar images that represent ``the same world state'' should map to the same representation --
and \emph{bijection} -- ability to map symbolic states to real-world images.
We need a bidirectional mapping where the symbolic representation captures the ``essence'' of the image, not merely the literal, raw pixel vector.

\begin{figure}[tbp]
 \centering
 \includegraphics[width=\linewidth]{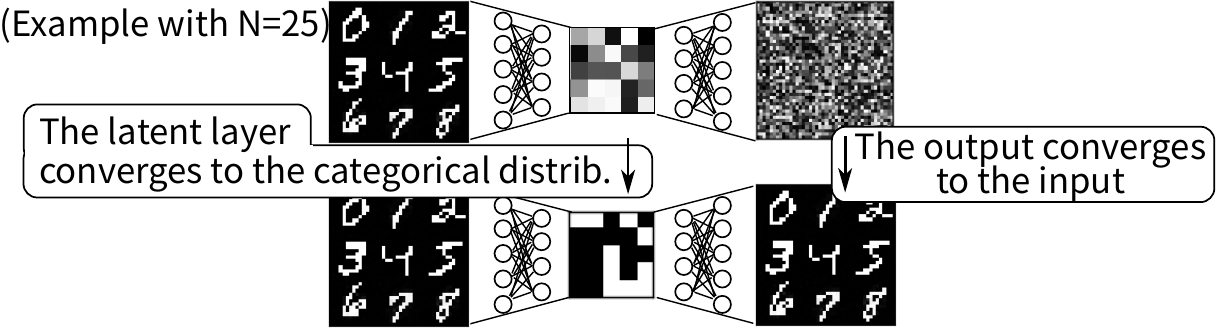}
 \caption{Step 1:
Train the State Autoencoder by
 minimizing the sum of the reconstruction loss and the variational loss of Gumbel-Softmax.
As the training continues, the output of the network converges to the input images.
Also, as the Gumbel-Softmax temperature $\tau$ decreases during training,
the latent values approach either 0 or 1.}
\label{fig:sae}
\end{figure}

The first technical contribution of this paper is the proposal of a SAE which is implemented as 
a Variational Autoencoder \cite{kingma2014semi} with a Gumbel-Softmax (GS) activation \cite{jang2016categorical} (\refig{fig:sae}).

A Variational Autoencoder (VAE) \cite{kingma2013auto} is a type of AE that forces the \emph{latent layer} (the most compressed layer in the AE) to follow a certain distribution (e.g., Gaussian).
Since the random distribution is not differentiable (BP is not applicable), VAEs use \emph{reparametrization tricks}, which decompose the target distribution into a differentiable and a purely random distribution (the latter does not require the gradient).
For example, the Gaussian $N(\sigma,\mu)$ is decomposed to $\mu+\sigma N(1,0)$, where $\mu,\sigma$ are learned.
In addition to the reconstruction loss, VAE should also minimize the variational loss (the difference between the learned and the target distributions) measured by, e.g.,  KL divergence.

Gumbel-Softmax (GS) is a recently proposed re\-para\-metri\-zation trick \cite{jang2016categorical} for categorical distribution.
It continuously approximates Gumbel-Max \cite{maddison2014sampling}, a method for drawing categorical samples.
Assume the output $z$ is a one-hot vector, e.g. if the domain is $D=\braces{a,b,c}$, $\brackets{0,1,0}$ represents ``b''.
The input is a class probability vector $\pi$, e.g. $\brackets{.1,.1,.8}$.
Gumbel-Max draws samples from $D$ following $\pi$:
 $z_i \equiv [ i = \arg \max_j (g_j+\log \pi_j)\ ?\ 1 : 0 ]$
where $g_j$ are i.i.d samples drawn from Gumbel$(0,1)$ \cite{gumbel1954statistical}.
Gumbel-Softmax approximates argmax with softmax to make it differentiable:
$z_i = \text{Softmax}((g_i+\log \pi_i)/\tau)$.
``Temperature'' $\tau$ controls the magnitude of approximation, which is annealed to 0 by a certain schedule.
The output of GS converges to a discrete one-hot vector when $\tau\approx 0$.

{\it Our key observation is that these categorical variables can be used directly as propositional symbols by a symbolic reasoning system, i.e., this gives a solution to the propositional symbol grounding in our architecture}.
In the SAE, we use GS in the latent layer. Its input is connected to the encoder network. The output is an $(N, M)$ matrix where $N$ is the number of categorical variables and $M$ is the number of categories.
We specify $M=2$, effectively obtaining $N$ propositional state variables. It is possible to specify different $M$ for each variable and represent the world using multi-valued representation as in SAS+ \cite{backstrom1995complexity}, but we use $M=2$ for all variables for simplicity.
This does not affect the expressiveness because bitstrings of sufficient length can represent arbitrary integers.

The trained SAE provides a bidirectional mapping between the raw inputs (subsymbolic representation) and their symbolic representations:
\begin{itemize} 
\setlength{\itemsep}{-0.3em}
\item $b=Encode(r)$ maps an image  $r$ to a boolean vector $b$.
\item $\tilde{r}=Decode(b)$ maps a boolean vector $b$ to an image $\tilde{r}$.
\end{itemize}
$Encode(r)$ maps raw input $r$ to a symbolic representation by feeding the raw input to the encoder network, extract the activation in the GS layer,
 and take the first row in the $N \times 2$ matrix, resulting in a binary vector of length $N$. Similarly, $Decode(b)$ maps a binary vector $b$  back to an image by concatenating $b$ and its complement $\bar{b}$ to obtain a $N \times 2$ matrix and feeding it to the decoder.
These are lossy compression/decompression functions, so in general, $\tilde{r}=Decode(Encode(r))$ may have an acceptable amount of errors from $r$ for visualization.

It is {\it not} sufficient to simply use traditional activation functions such as sigmoid or softmax and round the continuous activation values in the latent layer to obtain discrete 0/1 values.
In order to map the propositional states back to images,
we need a decoding network trained for 0/1 values.
A rounding-based scheme would be unable to restore the images because the decoder is not trained with inputs near 0/1 values.
Also, embedding the rounding operation as a layer of the network is infeasible because rounding is non-differentiable, precluding BP-based training of the network.


SAE implementation can easily and largely benefit from the progress in the image processing community.
We implemented SAE as a denoising autoencoder \cite{vincent2008extracting} to add noise robustness,
with some techniques which improve the accuracy (see Appendix \refsec{sec:SAE-detail}).

\section{AMA$_1$: Oracular PDDL Generator}
\label{sec:ama1-overview}

In AMA$_1$, our first AMA method, the output is a PDDL definition for a grounded unit-cost STRIPS planning problem.
AMA$_1$ is a trivial, \emph{oracular} strategy which generates a model based on \emph{all} transitions, i.e., 
$Tr$ contains image pairs representing all transitions that are possible in this domain, and
$\overline{Tr}$ contains all corresponding symbolic transitions.
The images are generated by an external, domain-specific image generator.
It is important to note that while $Tr$ for AMA$_1$ contains all transitions, the SAE is trained using only a subset of state images.
Although ideally an AMA component should induce a complete action model from a limited set of transitions,
AMA$_1$ is intended to demonstrate the overall feasibility of SAE-produced propositions and the overall \latentplanner architecture.

AMA$_1$ compiles $\overline{Tr}$ directly into a PDDL model as follows.
Each transition $\overline{tr}_i \in \overline{Tr}$ directly maps to an action $a_i$.
Each bit $b_j (1 \leq j \leq N)$ in boolean vectors $s_i$ and $t_i$ is mapped to propositions \texttt{(b$_j$-true)} and \texttt{(b$_j$-false)} when the encoded value is 1 and 0 (resp.). 
$s_i$ is directly used as the preconditions of action $a_i$.
The add/delete effects of action $i$ are computed by taking the bitwise difference between $s_i$ and $t_i$.
For example, when $b_j$ changes from 1 to 0, the effect compiles to \texttt{(and (b$_j$-false) (not (b$_j$-true)))}.
The initial and the goal states are similarly created by applying the SAE to the initial and goal images.

The PDDL instance output by AMA$_1$ can be solved by an off-the-shelf planner.
We use a modified version of Fast Downward \cite{Helmert2006} (see Appendix \refsec{sec:ama1-planner}).
\latentplanner inherits all of the search-related properties of the planner which is used. 
For example, if the planner is complete and optimal, \latentplanner will find an optimal plan for the given problem (if one exists), with respect to the portion of the state-space graph captured by the Action Model.


\subsection{Evaluating AMA$_1$ on Various Puzzles}

\label{sec:ama1-experiments}

We evaluated \latentplanner with AMA$_1$ on several puzzle domains.
Resulting plans are shown in \refigs{fig:mnist}{fig:mnist2}.
See Appendix \refsec{sec:domain-details} for further details of the network, training and inputs.

\textbf{MNIST 8-puzzle}
is an image-based version of the 8-puzzle, where tiles contain hand-written digits (0-9) from the  MNIST database \cite{lecun1998gradient}.
Valid moves in this domain swap the ``0'' tile  with a neighboring tile, i.e., the ``0'' serves as the ``blank'' tile in the classic 8-puzzle. 
The \textbf{Scrambled Photograph 8-puzzle (Mandrill, Spider)} cuts and scrambles real photographs, similar to the puzzles sold in stores).
These differ from the MNIST 8-puzzle in that ``tiles'' are \textit{not} cleanly separated by black regions
(we re-emphasize that \latentplanner has no built-in notion of square or movable region).
In \textbf{Towers of Hanoi (ToH)},
we generated the 4 disks instances.
4-disk ToH resulted in a 15-step optimal plan.
\textbf{LightsOut} is
a video game where a grid of lights is in some on/off configuration ($+$: On),
and pressing a light toggles its state as well as the states of its neighbors.
The goal is all lights Off.
Unlike previous puzzles, a single operator can flip 5/16 locations at once and
removes some ``objects'' (lights).
This demonstrates that \latentplanner is not limited to domains with highly local effects and static objects.
\textbf{Twisted LightsOut} distorts the original LightsOut game image by a swirl effect, 
showing that \latentplanner is not limited to handling rectangular ``objects''/regions.

\begin{figure}[tbp]
 \centering
 \includegraphics[width=\linewidth]{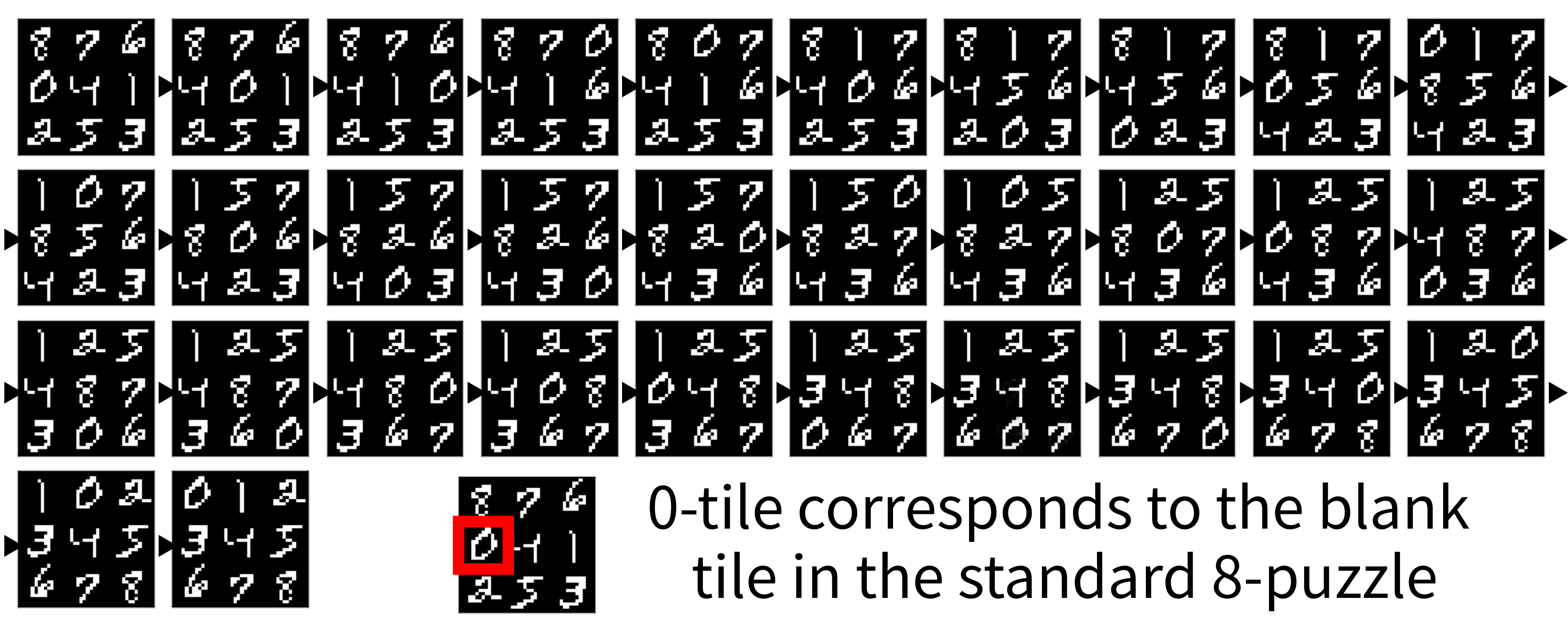}
 \includegraphics[width=\linewidth]{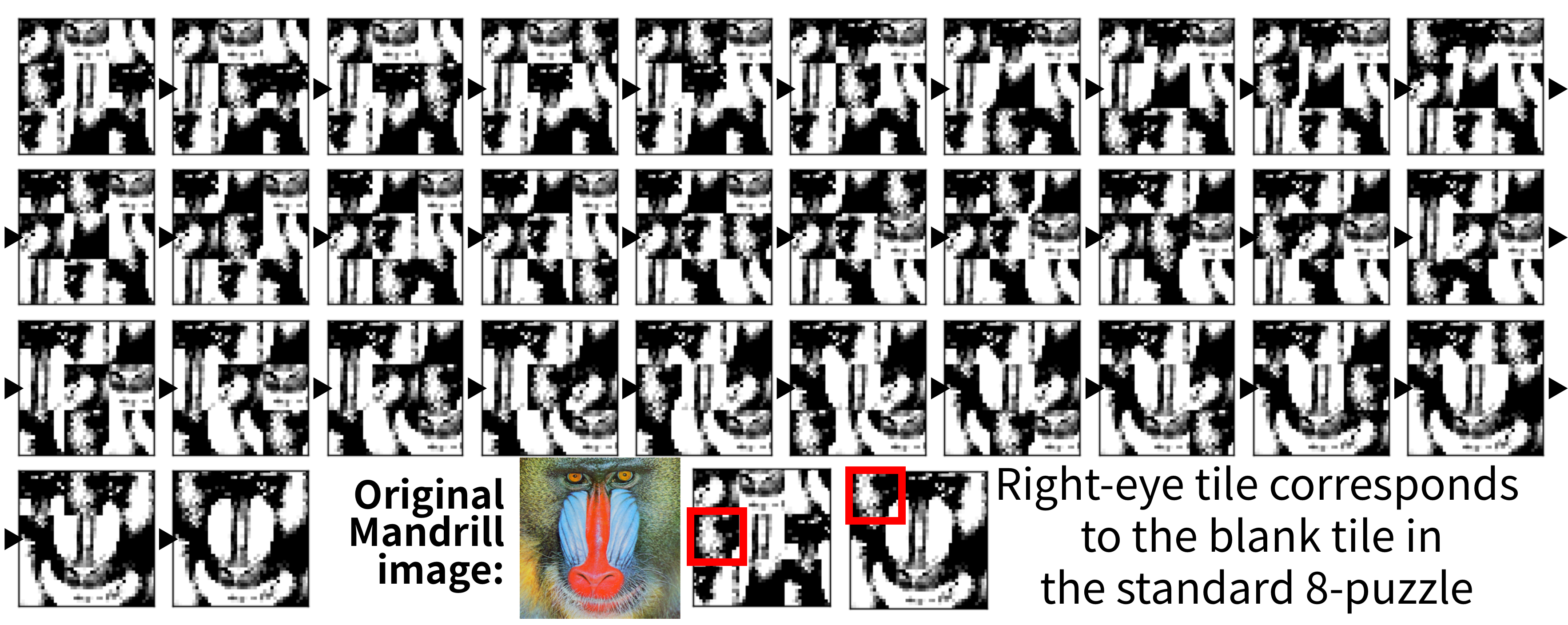}
 \includegraphics[width=\linewidth]{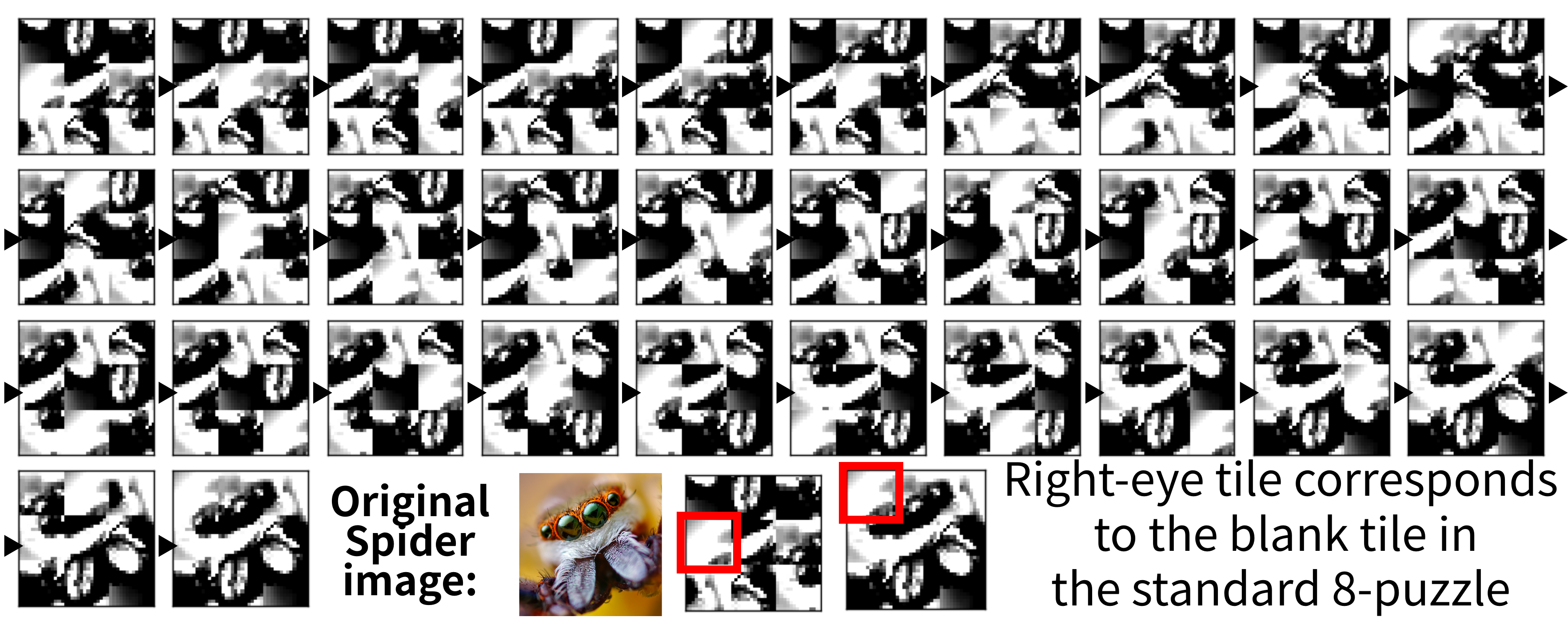}
 \caption{
Output of \latentplanner + AMA$_1$ solving the MNIST/Mandrill/Spider 8-puzzle instance
with the longest (31 steps) optimal plan (Reinefeld 1993).
This shows that \latentplanner finds an optimal solution
given a correct model by AMA$_1$ and an admissible search algorithm.
\latentplanner has no notion of ``slide'' or ``tiles'',
making MNIST, Mandrill and Spider entirely distinct domains.
SAEs are trained from scratch without knowledge transfer.
}
 \label{fig:mnist}
\end{figure}

\begin{figure}[tbp]
 \includegraphics[width=\linewidth]{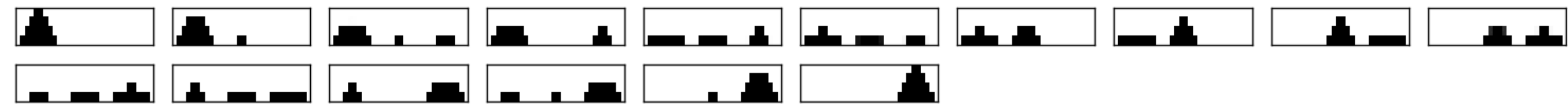}
 \includegraphics[width=\linewidth]{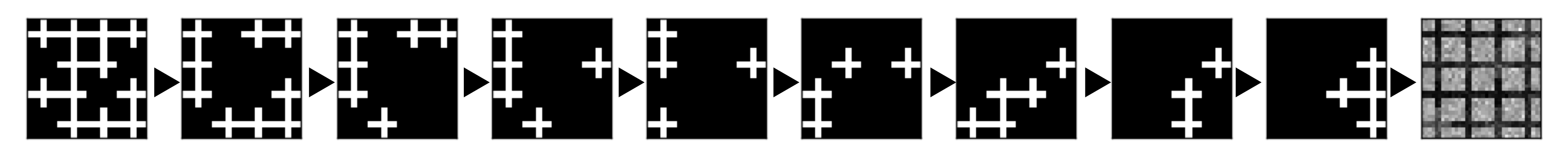}
 \includegraphics[width=\linewidth]{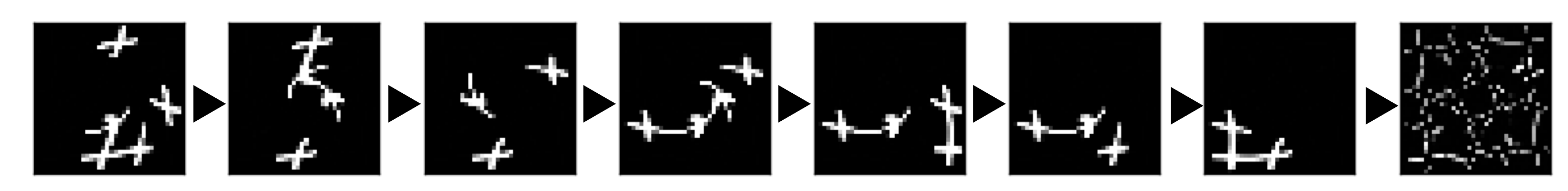}
 \caption{
(1) Output of solving ToH with 4 disks.
(2-3) Output of solving 4x4 LightsOut and Twisted LightsOut.
The blurs in the goal states are simply the noise that was normalized and enhanced by the plotting library.
}
 \label{fig:mnist2}
\end{figure}

\subsection{Robustness to Noisy Input}

\label{sec:noise-experiments}

\begin{figure}[tbp]
 \centering
 \includegraphics[width=\linewidth]{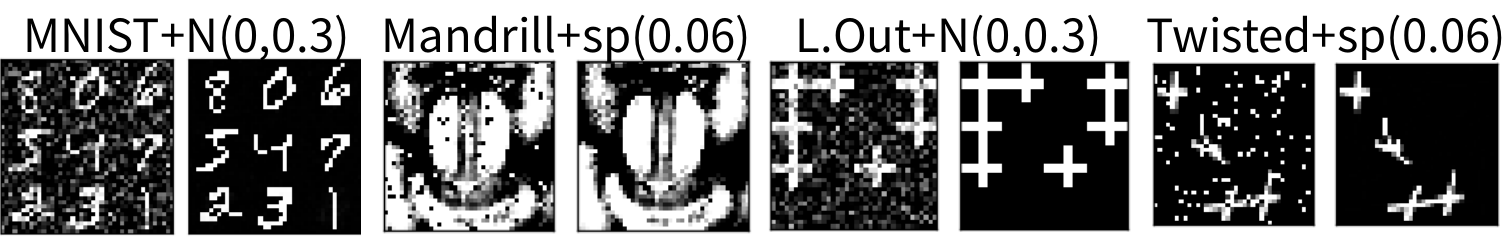}
 \caption{
SAE robustness vs noise:
  Corrupted initial state image $r$ and its reconstruction $Decode(Encode(r))$.
 Images are corrupted by Gaussian noise of $\sigma$ up to $0.3$ and by salt/pepper noise up to $p=0.06$.
 \latentplanner successfully solved the problems.
 The SAE maps the noisy image to the correct symbolic vector, finds a plan, then
 maps the plan back to the denoised images.
 }
  \label{fig:noise}
\end{figure}

\refig{fig:noise} demonstrates the robustness of the system vs. input noise.
We corrupted the initial/goal state inputs by adding Gaussian or salt/pepper noise.
The system is robust enough to successfully solve the problem
because of the Denoising AE \cite{vincent2008extracting}.

\section{AMA$_2$: Action Symbol Grounding}
\label{sec:ama2-overview}

\latentplanner + AMA$_1$ shows that (1) the SAE can robustly learn image $\leftrightarrow$ propositional vector mappings from examples, and that (2) if all valid image-image transitions (i.e., the entire state space) is given, \latentplanner can correctly generate optimal plans.
However, AMA$_1$ is clearly not practical due to the requirement that it uses the entire state space as input, and lacks the ability to learn/generalize an action model from a small subset of valid action transitions (image pairs).
Next, we propose AMA$_2$, a novel neural architecture  which jointly grounds the action symbols and acquires the action model from the subset of examples, in an unsupervised manner.

Acquiring a descriptive action model (e.g., PDDL) from a set of unlabeled propositional state transitions consists of three steps.
(Step 1) Identify the ``types'' of transitions, where each ``type'' is an identifiable, \emph{action symbol}.
For example, a  hand-coded ``slide-up-8-at-1-2'' in 8-puzzle is an example of action symbols, but note that an AMA system should ground anonymous symbols without human-provided labels.
While they are not lifted/parameterized, they still provide abstraction. For example, the same ``slide-up-8-at-1-2'' action, which slides the tile 8 at position $(x,y)=(1,2)$ upward, applies to many states (each state being a permutation of tiles 1-7).
(Step 2) Identify the preconditions and the effects of each action and store the information in an action model.
(Step 3) Represent the model in a modeling language (e.g., PDDL) as in \refig{8puzzle-pddl}.

Addressing this entire process is a daunting task. 
Existing AMA methods typically handle only Steps 2 and 3, skipping Step 1.
Without step 1, however, an agent lacks the ability to learn in an unknown environment where it does not know \emph{what is even possible}.
Note that even if the agent has the full knowledge of its low-level actuator capabilities, it does not know its own high-level capabilities e.g. sliding a tile.
Note that AMA$_1$ handles only Step 3, as providing all valid transitions is equivalent to skipping Step 1/2.

On the other hand, search on a state space graph in an unknown environment is \textit{feasible} even if Step 3 is missing.
PDDL provides two elements, a \emph{successor function} and its \emph{description}.
While ideally both are available, the description is not the \emph{essential} requirement.
The description may increase the explainability of the system in a language such as PDDL,
but such explainability may be lost anyway when the propositional symbols are identified by SAE, as the meanings of such propositions are unclear to humans (\refsec{sec:overview}).
The description is also useful for constructing the heuristic functions, but
the recent success of simulator-based planning \cite{frances2017purely}
shows that, in some application, efficient search is possible without action descriptions.
 
The new method, AMA$_2$, thus focuses on Steps 1 and 2.
It grounds the action symbols (Step 1) and finds a successor function that can be used for forward state space search (Step 2), but maintains its implicit representation.
AMA$_2$ comprises two networks: an \emph{Action Autoencoder} (AAE) and an \emph{Action Discriminator} (AD). The AAE jointly learns the action symbols and the action effects, and provides the ability to enumerate the candidates of the successors of a given state. The AD learns which transitions are valid, i.e. preconditions. Using the enumeration \& filtering approach, the AAE and the AD provides a successor function that returns a list of valid successors of the current state. Both networks are trained unsupervised, and operate in the symbolic latent space, i.e. both the input and output are SAE-generated bitvectors. This keeps the network small and easy to train.

\subsection{Action Autoencoder}

Consider a simple, linear search space with no branches.
In this case, grounding the action symbol is not necessary and
the AMA task reduces to predicting the next state $t$ from the current state $s$.
A NN $a'$ could be trained for a successor function $a(s)=t$, minimizing the loss $|t-a'(s)|$.
This applies to much of the work on scene prediction from videos such as \cite{srivastava2015unsupervised}. 


However, when the current state has multiple successors, as in planning problems, such a network cannot be applied.
One might consider training a separate NN for each action, but
(1) it is unknown how many types of transitions are available,
(2) the number of transitions depends on the current state, and
(3) it does not know which transition belongs to which action.
Although a single NN could learn a multi-modal distribution,
it lacks the ability to \emph{enumerate} the successors,
a crucial requirement for a search algorithm.

\begin{figure}[tbp]
 \centering
 \includegraphics[width=\linewidth]{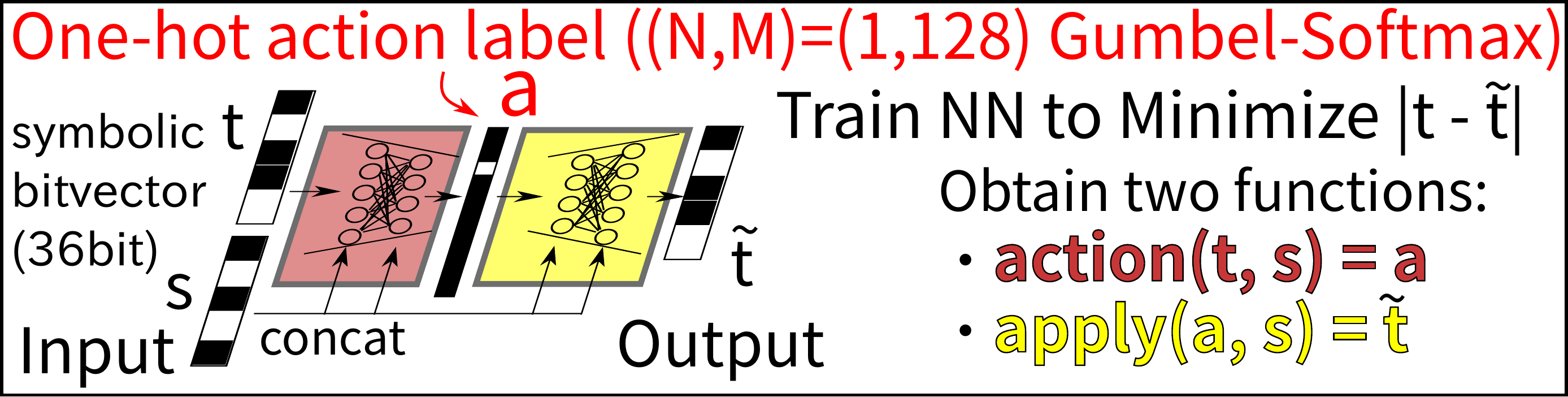}
 \caption{Action Autoencoder.}
 \label{fig:aae}
\end{figure}

To solve this, we propose an Action Autoencoder (AAE, \refig{fig:aae}). 
The key idea of AAE is to reformulate the transitions as $apply(a,s)=t$, which lifts the action symbol and makes it trainable,
and to realize that $s$ is the \emph{background information} of the state transition function.
The AAE has $s,t$ as inputs and reconstructs $t$ as $\tilde{t}$ whose error $|t-\tilde{t}|$ is minimized.
The main difference from a typical AE is:
(1) The latent layer is a Gumbel-Softmax one-hot vector indicating the \textbf{action label} $a$. 
(2) Every layer is concatenated with $s$.
The latter conditions the entire network by $s$,
which makes the 128 action labels (7bit) represent only the \emph{conditional information} (difference) necessary to ``reconstruct $t$ \emph{given} $s$'',
unlike typical AEs which encode the \emph{entire} information of the input.
%
As a result, the AAE learns the bidirectional mapping between $t$ and $a$, both conditioned by $s$:
\begin{itemize}
\setlength{\itemsep}{-0.3em}
 \item $Action(t,s)=a$ returns the action label from $t$.
 \item $Apply(a,s)=\tilde{t}$ applies $a$ to $s$ and returns a successor $\tilde{t}$. 
\end{itemize}

The number of labels serves as the upper bound on  the number of action symbols learned by the network.
Too few labels make AAE reconstruction loss fail to converge to zero.
After training, some labels may not be mapped to by any of the example transitions.
In the later phases of \latentplanner, these unused labels are ignored.
Since we obtain a limited number of action labels,
we can enumerate the candidates of the successor states of the given current state in constant time.
Without AAE, all $2^N$ states would be enumerated as the potential successors, which is clearly impractical.

\subsection{Action Discriminator}

\begin{figure}[tbp]
 \centering
 \includegraphics[height=0.7\paperheight]{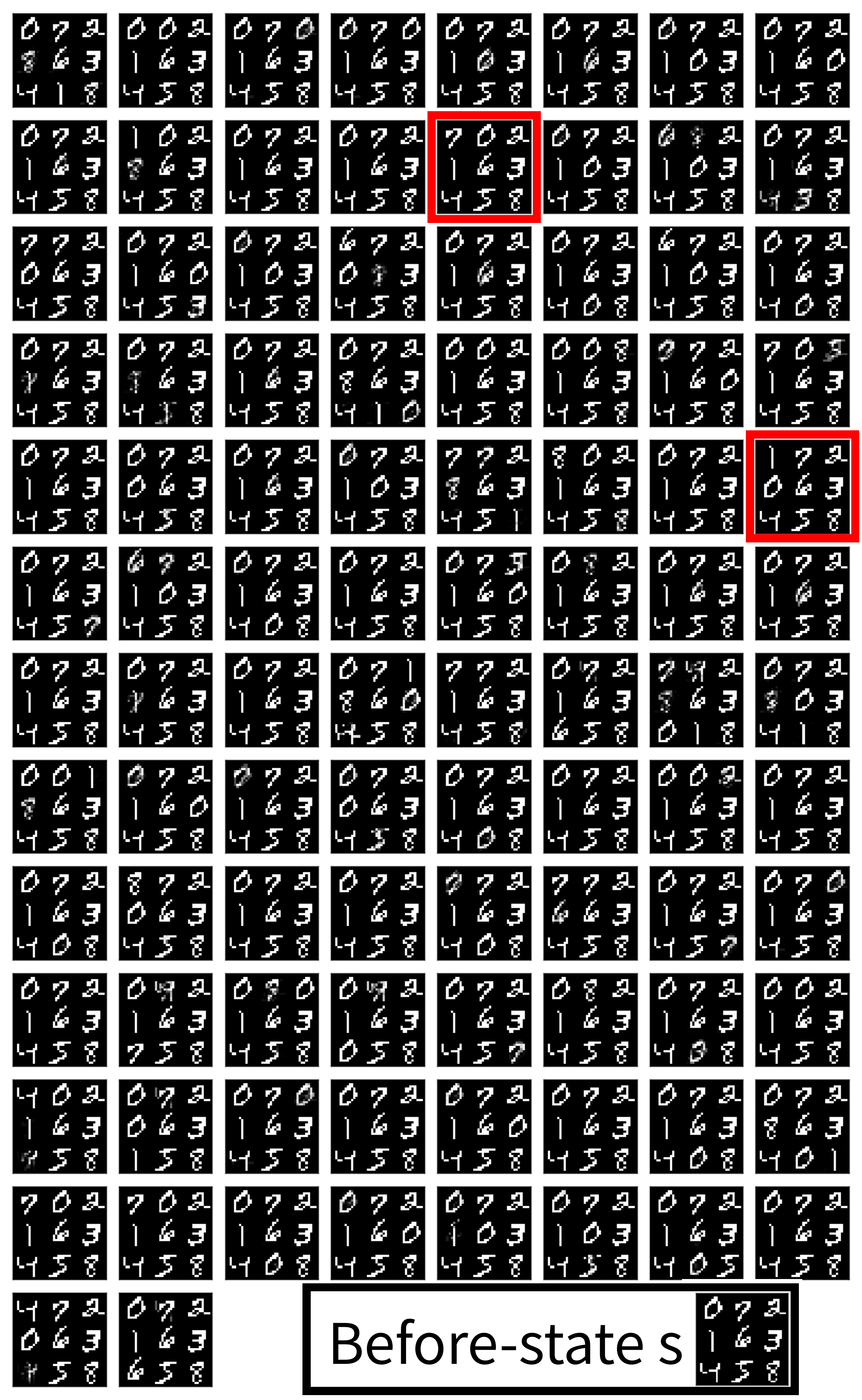}
\caption{The successors of a state $s$ (bottom-right),
  generated by applying all 98 actions identified by the AAE.
 A valid successor is marked by the red border.}
 \label{fig:aae-mixed}
\end{figure}

\begin{figure}[tbp]
 \centering
 \includegraphics[width=0.5\linewidth]{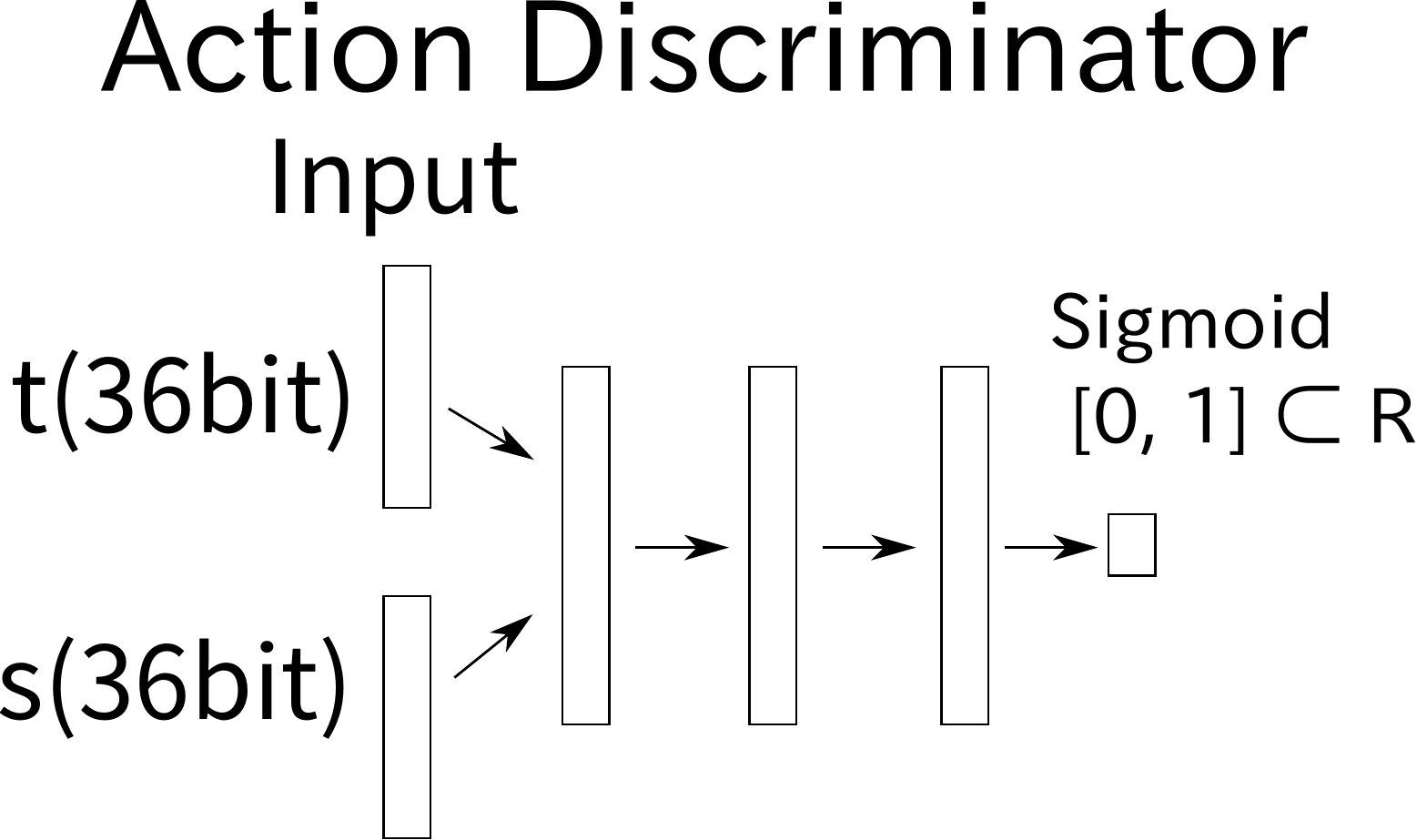}
 \caption{Action Discriminator.}
 \label{fig:ad}
\end{figure}

An AAE identifies the number of actions and learns the effects of actions, but does not address the applicability (preconditions) of actions.
Preconditions are necessary to avoid invalid moves (e.g. swapping 3 tiles at once) or invalid states (e.g. having duplicated tiles), as shown in \refig{fig:aae-mixed}.
Thus we need an \textit{Action Discriminator} (AD, \refig{fig:ad}) which learns the 0/1 mapping for each transition indicating whether it is valid, i.e., the ``preconditions''. This is a standard binary classification function which takes $s,t$ as inputs and returns a probability that $(s,t)$ is valid.

One technical problem in training the AD is that explicit \emph{invalid} transitions are unavailable.
This is not just a matter of insufficient data, but rather a fundamental constraint in an image-based system operating in the physical environment: Invalid transitions which violate the laws of physics (e.g. teleportation) are \emph{never} observed (because they never happens).
We then might consider ``imagining/generating'' the negative examples, as humans do in a thought experiment, but it is also impossible due to the lack of specification of \emph{what} is invalid.

To overcome this issue, we use the PU-Learning framework \cite{elkan2008learning}, which can learn a positive/negative classifier from the positive and \emph{mixed} examples that may contain both positive and negative examples.
We used $\overline{Tr}$ as the positive examples (they are all valid).
The mixed, i.e. possibly invalid, examples are generated by
applying each action $a$ (except unused ones) on each before-state $s$ in $\overline{Tr}$, and
removing the known positive examples from the generated pairs $(s,\tilde{t})$.

\subsection{PU-learning}
\label{sec:pu-learning}

The implementation of PU-learning is quite simple, following \cite{elkan2008learning}.
Given a positive ($p$) and a mixed ($m$) dataset,
 $p$ and $m$ are first arbitrarily divided into a training set ($p_1$ and $m_1$)
and validation set ($p_2$ and $m_2$), as usual.

Then, a binary classifier for $p_1$ (true) and $m_1$ (false) is trained.
As a result, we obtain a positive/mixed classifier $d_1(x)$ which is a function which returns a probability
that a data $x$ belongs to $p_1$.
After the training has finished,
the positive examples in the validation set ($p_2$) are classified, and
the  probability of $p_2$ belonging to $p_1$ are averaged to obtain a scalar $c = average(d_1(p_2))$.
As the final step, the true positive/negative classifier $d_2(x)$,
which is a function which returns a probability that a data $x$ is positive,
is defined as $d_2(x) = c \cdot d_1(x)$.

\subsection{State Discriminator}

\begin{figure}[tbp]
 \centering
 \includegraphics[width=0.5\linewidth]{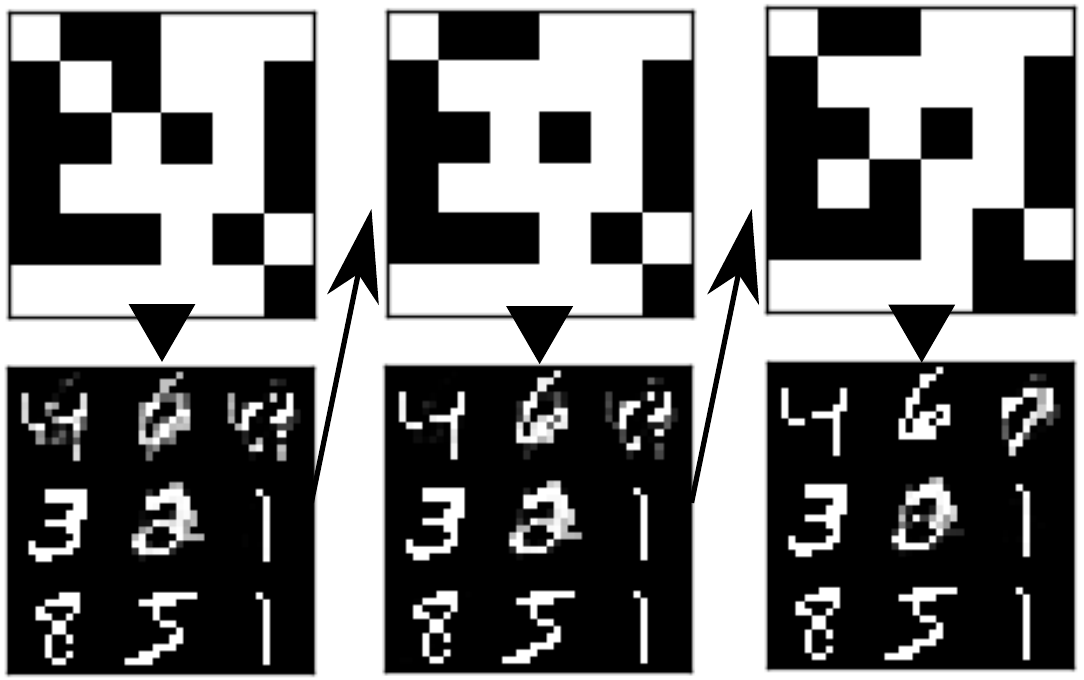}
 \caption{
(top, left) Random bit vector $\rho$,
(bottom, left) $Decode(\rho)$,
(top, middle)  $Encode(Decode(\rho))=\rho_2$,
(bottom, middle) $Decode(\rho_2)$,
(top, right)  $Encode(Decode(\rho_2))=\rho_3$,
(bottom, right) $Decode(\rho_3)$.
As more autoencoding is performed, images become less blurry,
 although they are still invalid (two 1-tiles).
}
\label{fig:random-bit}
\end{figure}

As a performance improvement, we also trained a State Discriminator (SD) which is a binary classifier for a single state $s$ and detects the invalid states, e.g. states with duplicated tiles in 8-puzzles. Again, we use PU-learning. Positive examples are the before/after states in $\overline{Tr}$ (all valid). Mixed examples are generated from the random bit vectors $\rho$ (may be invalid):
Many of the images $Decode$'ed from $\rho$ are blurry and do not represent autoencodable, meaningful real-world images.
However, when they are repeatedly encoded/decoded (\refig{fig:random-bit}), they converge to the clear, autoencodable invalid states because of the denoising AE \cite{vincent2008extracting}, and we used the results as the mixed examples.
If $Decode(\rho)$ results in a blurry image,
this ``blur'' is recognized as a noise and reduced in each autoencoding step,
finally resulting in a clean, reconstructable invalid image.
We use the SD to prune some mixed action examples for the AD training so that they contain only the valid successors.
This improves the AD accuracy significantly.

\subsection{Additional Pruning Methods}

Additionally, we have two more pruning methods:
The first one ensures that the SAE successfully reconstructs the successor state $t$, i.e. $t$ and $Encode(Decode(t))$ are identical.
The second one ensures that the AAE reconstructs $t$, i.e. $t$ and $Apply(Action(t,s),s)$ are identical.
Successors which failed to be reconstructed are removed from the consideration.

\subsection{Planning in \latentplanner using AMA$_2$}

In the case of AMA$_2$, we can not use an off-the-shelf PDDL-based planner because the action model is embedded in
the AAE, AD, and SD neural networks.
However, they allow us to implement a successor function which can be used in any symbolic, 
forward state space search algorithm such as \astar \cite{hart1968formal}.
The AAE generates the (potentially invalid) successors and
 the AD and SD filter the invalid states:
\begin{align*}
  Succ(s) &= \{t = apply(a,s) \; | \; a \in \braces{0\ldots 127} \setminus \textit{unused},\\
          & \qquad \land AD(s,t) \geq 0.5 \\
          & \qquad \land SD(t) \geq 0.5 \\
          & \qquad \land Encode(Decode(s)) \equiv s \\
          & \qquad \land Apply(Action(t,s),s) \equiv t \}
\end{align*}
We implemented A* in which states are latent-space (propositional) vectors, and the above $Succ$ function is used to generate successors of states.
A simple goal-count heuristic is used.
As the goal-count heuristics is inadmissible, the results could be suboptimal.
However, the purpose of implementing this planner is to see the feasibility of the action model.



\subsection{Evaluation}

\label{sec:ama2-experiments}

We evaluate the feasibility of the action symbols and the action models learned by AAE and AD.
We tested 8-puzzle (mnist, mandrill, spider), LightsOut (+ Twisted). 
We generated 100 instances for each domain and for each noise type (std, gaussian noise, salt/pepper noise)
 by  7-step (benchmark A)or  14-step (benchmark B) self-avoiding random walks from the goal state,
and evaluated the planner with the 180 sec.\ time limit.
We verified the resulting image plans with domain-specific validators.
\reftbl{tab:aae-results} shows that the \latentplanner achieves a high success rate.
The failures are due to timeouts
(the successor function requires many calls to the feedforward neural nets,
 resulting in a very slow node generation).

We next examine the accuracy of the AD and SD (\reftbl{tab:aae-results}).
We measured the type-1/2 errors for the valid and invalid transitions (AD) and states (SD).
Low errors show that our networks successfully learned the action models.

\begin{table}[tbp]
\centering
\begin{tabular}{|l|r|r|r|r|r|r||l|l|l|l|l|l|}
\hline
domain   & \multicolumn{3}{c|}{A:step=7}& \multicolumn{3}{c||}{B:step=14} &  \multicolumn{2}{c|}{SD error (\%)} &  \multicolumn{4}{c|}{AD error (in \%)} \\
                     & std & G   & s/p & std & G & s/p & type1   & type2   & type1  & type2  & 2/SD & 2/V  \\ \hline
MNIST                & 72  & 64  & 64  &6    &4  &3    & 0.09    & $<$0.01 & 1.55   & 14.9   & 6.15 & 6.20 \\ 
Mandrill             & 100 & 100 & 100 &9    &14 &14   & $<$0.01 & $<$0.01 & 1.10   & 16.6   & 2.93 & 2.94 \\ 
Spider               & 94  & 99  & 98  &29   &36 &38   & $<$0.01 & $<$0.01 & 1.22   & 17.7   & 4.97 & 4.91 \\ 
L. Out               & 100 & 99  & 100 &59   &60 &51   & $<$0.01 & N/A     & 0.03   & 1.64   & 1.64 & 1.64 \\
Twisted              & 96  & 65  & 98  &75   &68 &72   & $<$0.01 & N/A     & 0.02   & 1.82   & 1.82 & 1.82 \\
\hline
\end{tabular}
\caption{
AMA$_2$ results: (\textbf{left}) Number of solved instances out of 100 within 3 min. time limit.
The 2nd/3rd columns show the results when the input is corrupted by G(aussian) or s(alt)/p(epper) noise.
In benchmark A (created with 7-step random walks),
\latentplanner solved the majority of instances even under the input noise.
In the harder instances (benchmark B: 14-steps),
many instances were still solved. 
(\textbf{right}) Misclassification by SD and AD in \%, measured as:
(SD type-1) Generate all valid states and count the states misclassified as invalid.
(type-2) Generate reconstructable states, remove the valid states (w/ validator),
sample 30k states, and count the states misclassified as valid.
N/A means all reconstructable states were valid.
(AD type-1) Generate all valid transitions and count the number of misclassification.
(type-2) For 1000 randomly selected valid states, generate all successors, remove the valid transitions (w/ validator), then count the transitions misclassified as valid.
(2/SD, 2/V) Same as Type-2, but ignore the transitions whose successors are invalid according to SD or the validator.
Relatively large AD errors explain the increased number of failures in MNIST 8-puzzles.
}
\label{tab:aae-results}
\end{table}

\section{Related Work}
\label{sec:related}


Compared to the work by \citeauthor{KonidarisKL14} (\citeyear{KonidarisKL14}),
the inputs to \latentplanner are unstructured (42x42=1764-dimensional arrays for 8-puzzle);
each pixel does not carry a meaning and the boundary between ``identifiable entities'' is unknown.
Also, AMA$_2$ automatically grounds action symbols, while they rely on human-assigned symbols (\pddl{move, interact}).
Furthermore, they do not explicitly deal with robustness to noisy input, while we implemented SAE as a denoising AE.
However, effects/preconditions in AMA$_2$ is implicit in the network, and their approach could be utilized to extract PDDL from AAE/AD (future work).


%

There is a large body of work using NNs to directly solve combinatorial tasks,
starting with the well-known TSP solver \cite{hopfield1985neural}.
Neurosolver represents a search state as a node in NN 
and solved ToH \cite{bieszczad2015neurosolver}. 
However, they assume a symbolic input.

Previous work combining symbolic search and NNs embedded NNs {\it inside} a search
to provide the search control knowledge,
e.g., domain-specific heuristic functions for
the sliding-tile puzzle and Rubik's Cube \cite{ArfaeeZH11},
classical planning \cite{SatzgerK13},
or the game of Go \cite{alphago}.
Deep Reinforcement Learning (DRL) has solved complex problems,
including video games where it communicates to a simulator through images \cite[DQN]{dqn}.
In contrast, \latentplanner only requires a set of unlabeled image pairs (transitions), and 
does not require a reward function for unit-action-cost planning,
nor expert solution traces (AlphaGo),
nor a  simulator (DQN), nor predetermined action symbols (``hands'', control levers/buttons).
Extending \latentplanner to symbolic POMDP planning is an interesting avenue for future work.

A significant difference between \latentplanner and learning from observation (LfO) in the robotics literature \cite{ArgallCVB09} is that
\latentplanner is trained based on individual transitions
while LfO work is largely based on the longer sequence of transitions (e.g. videos)
and should identify the start/end of actions (\emph{action segmentation}).
Action segmentation would not be an issue in 
an implementation of autonomous \latentplanner-based agent
because it has the full control over its low-level actuators and
initiates/terminates its own action for the data collection.


\section{Discussion and Conclusion}

\label{sec:discussion}


We proposed  \latentplanner, an integrated architecture for learning and planning which,
given only a set of unlabeled images and no prior knowledge, generates a classical planning problem,
solves it with a symbolic planner,
and presents the plan as a human-comprehensible sequence of images.
%
We demonstrated its feasibility using image-based versions of planning/state-space-search problems (8-puzzle, Towers of Hanoi,  Lights Out).
Our key technical contributions are
(1) \emph{SAE, which leverages the Gumbel-Softmax to learn a bidirectional mapping between raw images and propositional symbols compatible to symbolic planners}.
On 8-puzzle, the ``gist'' of  42x42 training images are robustly compressed into propositions, capturing the essence of the images.
(2) \emph{AMA$_2$, which jointly grounds action symbols and learns the preconditions/effects.}
It identifies which transitions are ``same'' wrto the state changes and when they are allowed.

The only key assumptions about the input domain we make are that
(1) it is fully observable and deterministic and (2) NNs can learn from the available data.
Thus, we have shown that different domains can all be solved by the same system,
without modifying any code or the NN architecture.
In other words, \emph{\latentplanner is a domain-independent, image-based classical planner}.
To our knowledge, this is the first system which completely automatically constructs a logical representation
\emph{directly usable by a symbolic planner} from a set of unlabeled images for a diverse set of problems.

We demonstrated the feasibility of leveraging deep learning in order to enable 
symbolic planning using classical search algorithms such as A*, when only image pairs representing action start/end states are available,
and there is no simulator, no expert solution traces, and no reward function.
Although much work is required to determine the applicability and scalability of this approach,
we believe this is an important first step in bridging the gap between symbolic and subsymbolic reasoning and opens many avenues for future research.

\clearpage
\appendix

\section{State AutoEncoder}

\label{sec:SAE-detail}

All of the SAE networks used in the evaluation have the same network
topology for each domain, except the input layer which should fit the size of the input
images. They are implemented with TensorFlow and Keras libraries in under
5k lines of code.
We used a trivial, custom-made random grid search for automated tuning.
All layers except Gumbel-Softmax in the network are the very basic ones introduced in a standard tutorial.

The network uses a convolutional network in the encoder, and fc layers
in the decoder (\refig{fig:sae-detail}). The latent layer has 36 bits.
Input layer has the same dimension as the image size.
The network was trained using the Adam optimizer \cite{kingma2014adam}.
Learning rate (lr) starts at 0.001, and is decreased to 0.0001 at the half of the entire epoch.
In 8-puzzle domains, we used 150 epochs and batch-size 4000.
In LightsOut domains, we used 100 epochs and batch-size 2000, due to the larger size of the image.
In Hanoi, we used a channel size of 12 instead of 16 for convolutions, dropout 0.6, and batch-size 500.
Training takes about 15 minutes on a single NVIDIA GTX-1070.

\begin{figure}[htb]
\centering
\begin{tabular}{|l|}
 Input($input$),\\
 GaussianNoise(0.4),\\
 conv(3,3,16), tanh, bn, dropout(0.4),\\
 conv(3,3,16), tanh, bn, dropout(0.4),\\
 fc(72), reshape(36x2), GumbelSoftmax,\\
 fc(1000), relu, bn, dropout(0.4),\\
 fc(1000), relu, bn, dropout(0.4),\\
 fc($input$), sigmoid.
\end{tabular}
\caption{SAE implementation.
 Here, fc = fully connected layer, conv = convolutional layer, 
relu = Rectified Linear Unit,
bn = Batch Normalization, 
and tensors are reshaped accordingly.}
\label{fig:sae-detail}
\end{figure}


In all experiments, 
the annealing schedule of Gumbel-Softmax is $\tau \leftarrow \max (0.7, \tau_0\exp(-rt))$ where
 $t$ is the current training epoch, $\tau_0$ is the initial temperature and $r$ is an annealing ratio.
We chose $\tau_0,r$ so that $\tau = 5.0$ when the training starts and $\tau = 0.7$ when the training finishes.
The above schedule is similar to the original schedule in  \citeauthor{jang2016categorical} (\citeyear{jang2016categorical}).

\subsection{State Augmentation}

As mentioned in the paper, the number of bits should be larger than the minimum encoding length $\log_2 |S|$
of the entire state space $S$.
36 bits in the latent layer sufficiently covers
the total number of states in any of the problems that are used in the experiments.

However, excessive latent space capacity (number of bits) is also harmful.
Due to the nature of Gumbel-Softmax, which uses Gumbel random distribution,
excessive number of bits results in meaningless bits that does not affect the decoder output.
These bits act like purely random variables and cause multiple symbolic states to represent the same image.
This causes an undesirable behavior in the latent space,
since it could make the search graph disconnected.

One way to obtain a connected search graph under this condition is
what we call \emph{state augmentation},
which encodes the same image several times and simply sample the bitvectors for an image.
This technique is used in the Towers of Hanoi (ToH) AMA$_1$ experiments, as ToH has the small search space.

In general, there is a tradeoff: The larger the latent space capacity, the easier it is to train the SAE,
but the latent space becomes more stochastic.
Thus, it is desirable to reduce the latent capacity with further engineering,
while trying to connect the search graph with sampling.


\section{Action AutoEncoder}

\begin{figure}[htb]
\centering
\begin{tabular}{|l|}
 Input(36),                                                  \\
 concatenate(s), fc(400), relu, bn, dropbout(0.4),           \\
 concatenate(s), fc(400), relu, bn, dropbout(0.4),           \\
 concatenate(s), fc(128), reshape(1x128), GumbelSoftmax,     \\
 concatenate(s), fc(400), relu, bn, dropbout(0.4),           \\
 concatenate(s), fc(400), relu, bn, dropbout(0.4),           \\
 fc(36), sigmoid
\end{tabular}
\caption{AAE implementation. Here, fc = fully connected layer, bn = Batch Normalization,
and tensors are reshaped accordingly.}
\label{fig:aae-detail}
\end{figure}

AAE consists of the layers as shown in \refig{fig:aae-detail}.
The input takes the successor state $t$ (36bit) and concatenate(s) concatenate the input with the before-state $s$ (36bit).
The output of the network is $\tilde{t}$, a 36bit reconstruction of $t$.
The network was trained with lr:0.001, Adam, batch size 2000, 1000 epochs,
to minimize the reconstruction loss $|t-\tilde{t}|$ in terms of binary cross-entropy.
Training takes about 5 min.

In all experiments, the annealing schedule of Gumbel-Softmax is $\tau \leftarrow \max (0.1, \tau_0\exp(-rt))$.
We chose $\tau_0, r$ so that $\tau = 5.0$ and $\tau = 0.1$ when the training finishes.

\section{Action Discriminator (AD)}

The Action Discriminator uses PU-learning framework \cite{elkan2008learning}
to learn a positive/negative binary classifier from a positive/mixed
dataset.

We first concatenate $s$ and $t$, resulting in a set of 72 bit binary vectors.
We prepare a vector whose length is the same as the number of data, and assign 1 to the positive data,
and 0 to the mixed data.
$p$ and $m$ are concatenated, shuffled, and divided into training set (90\%) and the validation set (10\%).

To classify $p_1$ and $m_1$, we trained several networks shown in \refig{fig:ad-detail} and
chose the one which achieved the best accuracy.
The network is trained using Adam, lr:0.001, 3000 epochs with early stopping, batch size 1000,
using binary cross-entropy loss. Each training takes 2-10 minutes depending on the domain.

\begin{figure}[htb]
\centering
\begin{tabular}{|l|}
 Input(72), \\
 \ [bn, fc(300), relu, dropout($X$)] $\times Y$, \\
 fc(1), sigmoid.
\end{tabular}
\caption{AD implementation. It has metaparameters $X,Y$, where
$X\in \braces{0.5,0.8}$, $Y\in \braces{1,2}$, resulting in 4 configurations in total.
Depending on the value of $Y$, it becomes a single-layer or a two-layer perceptron.
}
\label{fig:ad-detail}
\end{figure}

\section{State Discriminator}

The State Discriminator also uses the PU-learning framework \cite{elkan2008learning}.
The dataset is prepared as described in the paper, and divided into training and validation set (90\% and 10\%).
We use the following single layer perceptron:
[Input(36), bn, fc(50), relu, dropout(0.8), fc(1), sigmoid].
The network is trained using Adam, lr:0.0001, 3000 epochs with early stopping, batch size 1000,
using binary cross-entropy loss.

\section{Domain Details}

\label{sec:domain-details}

\subsection{MNIST 8-puzzle}

This is an image-based version of the 8-puzzle, where tiles contain
hand-written digits (0-9) from the MNIST database
\cite{lecun1998gradient}. Each digit is shrunk to 14x14 pixels, so each
state of the puzzle is a 42x42 image.  Valid moves in this domain swap
the ``0'' tile with a neighboring tile, i.e., the ``0'' serves as the
``blank'' tile in the classic 8-puzzle.  The entire state space consists
of 362880 states ($9!$) and 967680 actions.  From any specific goal state, the reachable
number of states is 181440 ($9!/2$).  Note that the same image is used
for each digit in all states, e.g., the tile for the ``1'' digit is the
same image in all states.

We provide 20000 random transition images as $Tr$.
This contains 2x20000 images including the duplicates.
SAE learns from these 40000 images.
Next, SAE generates latent vectors of 2x20000 images, then use them as the input to AAE, AD and SD.
In all cases training:validation ratio 9:1 is maintained
 (i.e. only 36000 images and 18000 transitions are used for training).

\subsection{Mandrill, Spider 8-Puzzle}

These are 8-puzzles generated by cutting and scrambling real photographs
(similar to sliding tile puzzle toys sold in stores). We used the
``Mandrill'' and ``Spider'' images, two of the standard benchmark in the image processing
literature.  The image was first converted to greyscale and then
histogram-normalization and contrast enhancement was applied.
The same number of transitions as in the MNIST-8puzzle experiments are used.

\subsection{LightsOut}

A video game where a grid of lights is in some on/off configuration ($+$: On),
and pressing a light toggles its state (On/Off) as well as the state of all of its neighbors.
The goal is all lights Off.
(cf. \url{https://en.wikipedia.org/wiki/Lights_Out_(game)})
Unlike the 8-puzzle where each move affects only two adjacent tiles, a single operator in 4x4 LightsOut  can simultaneously flip 5/16 locations.
Also, unlike 8-puzzle and ToH, the LightsOut game allows some ``objects'' (lights) to disappear.
This demonstrates that \latentplanner is not limited to domains with highly local effects and static objects.

The image dimension is 36x36 and the size of each button ($+$ button) is 9x9.
4x4 LightsOut has $2^{16}=65536$ states and $16\times 2^{16}=1048576$ transitions.
Similar to the 8-puzzle instances, we used 20000 transitions.
Training:validation ratio 9:1 is maintained (i.e. only 36000 images and 18000 transitions are used for training).

\subsection{Twisted LightsOut}

The images have the same structure as LightsOut, but 
we additionally applied a swirl effect available in scikit-image package.
The effect is applied to the center, with strength=3, linear interpolation, 
and radius equal to 0.75 times the dimension of the image.

The image dimension is 36x36. 
Before the swirl effect is applied,
the size of each button ($+$ button) was 9x9.

\subsection{Towers of Hanoi}

Disks of various sizes must be moved from one peg to another, with the
constraint that a larger disk can never be placed on top of a
smaller disk.
Each input image has a dimension of $16\times 60$ (resp.),
where each disk is presented as a 4px line segment.

Due to the smaller state space ($3^d$ states for $d$ disks: 81 states, 240 transitions for 4 disks)
compared to the other domains tested in this paper,
we used 90\% of states as the training examples in AMA$_1$ experiments,
and verified on the 10\% validation set that the network is generalizing.

We also applied the state augmentation technique described in \refsec{sec:SAE-detail},
as the detrimental effect of excessive number of bits in the latent space becomes more obvious in this domain.

\section{Planner details}

\subsection{Planner in AMA$_1$ experiments}
\label{sec:ama1-planner}

In the AMA$_1$ experiments (\refsec{sec:ama1-experiments}),
we found that the invariant detection
routines in the Fast Downward PDDL to SAS translator (translate.py)
became a bottleneck.
This is because the PDDL represent individual transitions as ground actions, whose number is very large.
In order to run the experiments in \refsec{sec:ama1-experiments} 
,
we wrote a trivial, replacement PDDL to SAS converter without the invariant detection.
Still, each experiment may require more than 7GB memory and 4 hours on a Xeon E6-2676 CPU.
Most of the runtime was spent on the preprocessing, and the search takes only a few seconds.

\subsection{Planner in AMA$_2$ experiments}

In the AMA$_2$ experiments (\refsec{sec:ama2-experiments}), we implemented a trivial A* planner in python.
Although this implementation could be hugely inefficient compared to the traditional native-complied solvers,
the performance is not our concern.
In fact, the most time-consuming step is the generation and the filtering of the successor states using AAE, AD etc.,
and the low-level implementation detail is not the bottleneck.

The goal-count heuristics is based on the bitwise difference between the
latent representation of the goal image and the current state.

\section{Statement on Reproducibility}

To facilitate reproducibility of the experiments, the entire source code of the system and the pre-trained network weights
 will be made public on Github (\url{https://github.com/guicho271828/latplan}).

\fontsize{9.5pt}{10.5pt}
\selectfont
 

\bibliographystyle{named}

\end{document}